# Identifying Slug Formation in Oil Well Pipelines: A Use Case from Industrial Analytics

Abhishek Patange
ABB Global Industries and Services Pvt. Ltd.
Bengaluru, Karnataka, India
abhishek.patange@in.abb.com

Sharat Chidambaran
ABB Global Industries and Services Pvt. Ltd.
Bengaluru, Karnataka, India
sharat.chidambaran@in.abb.com

Prabhat Shankar
ABB Global Industries and Services Pvt. Ltd.
Bengaluru, Karnataka, India
prabhat.shankar@in.abb.com

Manjunath G.B.
ABB Global Industries and Services Pvt. Ltd.
Bengaluru, Karnataka, India
manjunath.gb@in.abb.com

Anindya Chatterjee
ABB Global Industries and Services Pvt. Ltd.
Bengaluru, Karnataka, India
anindya.chatterjee@in.abb.com

**Abstract**
Slug formation in oil and gas pipelines poses significant challenges to operational safety and efficiency, yet existing detection approaches are often offline, require domain expertise, and lack real-time interpretability. We present an interactive application that enables end-to-end data-driven slug detection through a compact and user-friendly interface. The system integrates (i) data exploration and labeling, (ii) configurable model training and evaluation with multiple classifiers, (iii) visualization of classification results with time-series overlays, and (iv) a real-time inference module that generates persistence-based alerts when slug events are detected. The demo supports seamless workflows from labeled CSV uploads to live inference on unseen datasets, making it lightweight, portable, and easily deployable. By combining domain-relevant analytics with novel UI/UX features such as snapshot persistence, visual labeling, and real-time alerting, our tool adds significant dissemination value as both a research prototype and a practical industrial application. The demo showcases how interactive human-in-the-loop ML systems can bridge the gap between data science methods and real-world decision-making in critical process industries, with broader applicability to time-series fault diagnosis tasks beyond oil and gas.

**Keywords**
Slug detection; Oil well pipelines; Industrial analytics; Machine learning; Predictive maintenance

## 1. Introduction

Slugging in oil and gas pipelines is a critical flow assurance problem with direct implications for safety, equipment reliability, and production efficiency. Pressure surges caused by slug formation can damage downstream components, interrupt steady flow, and force unplanned shutdowns that are both costly and hazardous. The early detection and characterization of slugging, therefore, remains a central requirement for stable pipeline operations.

Research efforts to mitigate this phenomenon generally fall into two categories: passive and active control. Passive control involves physical design modifications–such as altering pipeline geometry or installing slug catchers–to minimize slug formation. While effective under fixed conditions, these methods lack flexibility for dynamic production environments. Active control, on the other hand, employs real-time monitoring and feedback mechanisms to stabilize flow through operational adjustments, typically via valves or choke control. Researchers have modeled slug dynamics using mass balance equations or partial differential equations (PDEs) when reliable liquid and gas mass flow data are available. In their absence, estimation models are used to infer these flow rates from measurable parameters, enabling control even with limited sensing.

Despite progress in model-based control, rule- and threshold-based detection methods remain common in practice due to their simplicity. However, these methods are highly sensitive to sensor noise, require frequent recalibration, and fail to adapt under changing conditions. More recently, machine learning (ML) techniques have emerged as promising alternatives, capable of capturing complex, nonlinear relationships among process variables. Yet, most ML-based approaches operate offline, offering post-hoc analysis rather than real-time insight–limiting their effectiveness in operational decision-making. Bridging this gap through data-driven, real-time active control represents a key opportunity for advancing slug management, combining the adaptability of ML with the responsiveness of control-based frameworks. To contextualize this challenge, Figure 1 illustrates an offshore production process, where multiphase flow and complex pipeline topography often give rise to slugging phenomena.

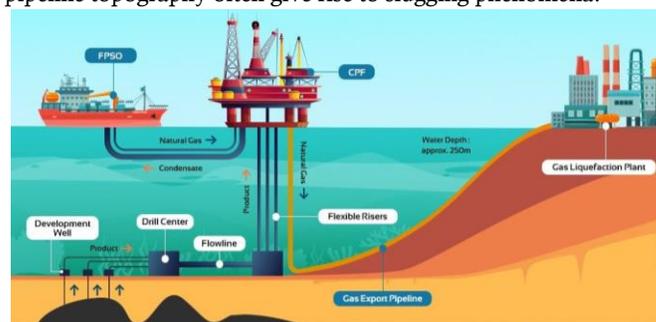

**Figure 1.** Offshore production process

Prior efforts show that both sensing and analytical advances have contributed to slug detection, a few of them are reviewed herein. Capacitance-based sensing has been widely studied for multiphase flow monitoring. Al-Alweet et al. (2020) demonstrated that electrode orientation substantially affects sensitivity and further proposed frequency–statistical mappings (variance, skewness, kurtosis, peak width, total power) for flow regime identification. However, two-electrode schemes are limited by low spatial resolution and poor void-fraction accuracy, while multi-electrode tomography, though richer, adds cost, calibration effort, and reconstruction noise. Distributed fiber-optic methods extend monitoring to long-distance pipelines. Ali et al. (2024) applied distributed acoustic sensing to quantify slug frequency, velocity, and length, validated against high-speed video. While promising for large-scale coverage, these methods are heavily signal-processing oriented and do not provide interactive, operator-centric analytics. At microfluidic scales, Gagliano et al. (2020) designed low-cost, real-time slug detection platforms based on optical sensing, cross-correlation, and GUI-based visualization. Similarly, Moscato et al. (2025) proposed an on-chip optical monitoring method for estimating slug-flow velocity in microchannels, validating against analytical models and CFD simulations. While effective for high-frequency characterization in Lab-on-a-Chip systems, such microfluidic approaches remain confined to controlled experimental environments. Both highlight the potential of compact, real-time tools, but remain tied to controlled laboratory setups and lack generalizability to industrial pipelines. Ran *et al.* (2025) introduced an AI-based framework for automated flow pattern classification using simple capacitance sensors, benchmarking multiple machine learning and deep learning models. However, their study emphasized algorithmic benchmarking rather than interactive, operator-facing workflows, leaving open the question of how such models can be integrated into practical decision-support systems. Mohmmed *et al.* (2016) developed an image processing technique combining foreground estimation, blob analysis, and morphological operations to automatically compute slug translational velocity and length from high-speed video. While the method achieved accuracy within 7% compared with reference measurements, its reliance on specialized imaging setups restricts practical deployment in industrial pipelines. Pedersen et al. (2015) reviewed slug modeling and control strategies, concluding that practical deployment is still hampered by incomplete plant knowledge, time-delay constraints, and calibration challenges. Thus, broader control literature emphasizes why interactive solutions are needed.

Overall, these studies demonstrate clear progress in sensing modalities and data-driven modeling for slug detection yet also reveal a critical gap–the absence of interactive and deployable system that enable domain experts to explore data, label events, train models, and receive interpretable alerts in real-time. This motivates our demonstration, designed to bridge that gap through an operator-centered, human-in-the-loop workflow.

## 2. Key concepts used in slug identification workflow

The demonstration tool follows a structured flow: a dataset is uploaded, labels are explored, models are trained on selected intervals, evaluation is visualized on the timeline, and inference is performed on unseen data with alerts generated in real time. Several key calculations that distinguish the workflow and enhance its practical value are discussed herein.

### 2.1 Time-aware data split

The demonstration system formalizes slug detection as a supervised classification problem on a multivariate time series. The goal here is to respect chronology and operating drift so evaluation reflects real deployment. Let the dataset be represented as $(x_t, y_t)$, where $x_t \in \mathbb{R}^d$ are sensor measurements and $y_t \in \{0,1\}$ indicates label whether slugging (*slug*=1) occurs at time $t$. Data are sampled every $\Delta_s$ minutes, with $\Delta_s = 5$ in our case. Users select disjoint train and test intervals $\mathcal{T}_{train}$ and $\mathcal{T}_{test}$, ensuring temporal integrity and avoiding look-ahead leakage. The datasets are defined as,

$$\mathcal{D}_{train} = \{ (x_t, y_t): t \in \mathcal{T}_{train} \} \,; \mathcal{D}_{test} = \{ (x_t, y_t): t \in \mathcal{T}_{test} \}$$

This chronological splitting is critical for evaluating how the models generalize under realistic operating drift. Process plants naturally drift across shifts, wells, seasons, maintenance windows etc., and random shuffling of data can overestimate performance.

### 2.2 Learners

For slug recognition, we employ three widely used learners: J48-Decision Tree, Random Forest, and XGBoost. Decision trees use the Gini index as a measure of node impurity to evaluate potential splits during training. At each node, the algorithm examines all possible feature thresholds and calculates the Gini impurity for the resulting child nodes.

Mathematically,

$$\Delta Gini = \left\{ \left[ 1 - \sum_{\{k=0\}}^{1} p_k^2 \right] - \left[ \sum_{\{c \in \{L,R\}\}} \frac{|S_c|}{|S|} Gini(S_c) \right] \right\}$$

Random forests mitigate variance by aggregating an ensemble of such trees, while XGBoost optimizes a regularized logistic objective using second-order updates. To address class imbalance, weights $w_1$ and $w_0$ is assigned to the positive (slug) and negative (non-slug) classes in the training loss, ensuring that rare slug events are not underrepresented.

### 2.3 Probability scoring & threshold selection

The goal here is to produce probabilities that reflect true risk and choose a decision threshold consistent with operational costs. Each trained model produces a probability score for each time $t$ as,

$$p_t = P(y_t = 1 \mid x_t)$$

These probabilities are calibrated using Platt scaling to improve reliability. Binary predictions are then obtained by thresholding:

$$\hat{y}_t = 1\{ p_t \geq \tau \}.$$

Threshold $\tau$ is tuned either to maximize the $F_\beta$ score,

$$F_\beta = \frac{(1+\beta^2) \cdot Precision \cdot Recall}{\beta^2 \cdot Precision + Recall}$$

$$Precision = \frac{TP}{TP+FP}, Recall = \frac{TP}{TP+FN}$$

with $\beta > 1$ prioritizing recall in safety-critical settings, or by maximizing Youden's index on the ROC curve,

$$J(\tau) = TPR(\tau) + TNR(\tau) - 1$$

### 2.4 Event-level visualization on the timeline

A notable feature of the workflow is the visualization of predictions on the time axis aiming to make performance interpretable where it happens—on the process timeline. Parameter value corresponding to each timestamp is mapped to one of four categories: true positive (TP), false positive (FP), true negative (TN), or false negative (FN). Formally,

$$c_t = \begin{cases} TP & y_t = 1, \quad \hat{y}_t = 1 \\ FP & y_t = 0, \quad \hat{y}_t = 1 \\ TN & y_t = 0, \quad \hat{y}_t = 0 \\ FN & y_t = 1, \quad \hat{y}_t = 0 \end{cases}$$

Practically contextual errors, such as false positives during start-ups or false negatives during pressure transients would be identified easily. It also highlights error clustering, where repeated false positives may signal that the decision threshold is too low or that sensor noise is driving spurious detections. This mapping enables users to see where misclassifications occur relative to operating conditions, providing deeper insight than aggregate scores alone.

Beyond pointwise metrics, many plants reason in terms of episodes, i.e., contiguous slug intervals. To evaluate this, event-based metrics are computed using Intersection-over-Union (IoU). For a predicted episode Ê and a true episode $E$, IoU is defined as:

$$IoU = \frac{|\hat{E} \cap E|}{|\hat{E} \cup E|}$$

A predicted event is considered a true detection if $IoU \geq \theta$, with $\theta$ typically set to 0.5. This event-level evaluation complements pointwise scores and aligns more closely with operator perception of flow instabilities, ensuring that the system captures practically meaningful events rather than isolated data-point classifications.

### 2.5 Persistence-based alerting

To suppress spurious alarms, inference employs persistence-based alerting. A run-length counter $r_t$ is updated as

$$r_t = \begin{cases} r_{\{t-1\}} + 1 & \hat{y}_t = 1, \\ 0 & \hat{y}_t = 0, \end{cases} \quad r_0 = 0$$

and an alert is raised only if $r_t \geq \kappa$,
where $\kappa$ corresponds to a minimum sustained slug duration

$$\kappa = \left(\frac{\Delta_{min}}{\Delta_s}\right)$$

thus, $A_t = 1\{r_t \geq \kappa\}$

This rule ensures that alerts correspond to genuine slug episodes rather than transient fluctuations. The goal of this mechanism is to avoid alarm chatter and only alert on sustained slug states. The run-length counter increments with each consecutive positive detection and resets when a negative occurs. An alert is triggered once the counter exceeds $\kappa$. Choosing $\Delta_{min}$ involves a trade-off:

- Short $\Delta_{min}$ allows faster detection but increases the risk of false positives.
- Long $\Delta_{min}$ reduces false positives but increases detection latency, approximately $\frac{(\kappa-1)}{\Delta_s}$
- A typical starting point is 10–20 minutes for 5-minute sampling, corresponding to $\kappa = 2 - 4$.
- To further stabilize alerts, hysteresis or a cool-down mechanism may be used.
- After an alert is raised, new alerts can be suppressed for $C$ samples to avoid flapping; the system resets once predictions return to zero for a minimum gap.
- Alternative filtering strategies can also be applied. One option is a sliding-window majority vote, where an alert is raised if:

$$\sum_{i=0}^{L-1} \hat{y}_{t-i} \geq m$$

with the ratio $\frac{m}{L}$ tuned for sensitivity.

- Another option is a two-threshold scheme: the system enters an alert state when probabilities exceed a high threshold τ_hi, and only exits when probabilities fall below a lower threshold $\tau_{lo}$ (with $\tau_{lo} < \tau_{hi}$).
- This hysteresis-based approach prevents oscillations around a single threshold.

### 2.6 Confidence Traces and Decision Rationale

Confidence trace visualization is introduced to communicate not only whether an alert was raised, but also the certainty associated with that decision. In this design, the probability trajectory $p_t$ is plotted alongside key process parameters. This joint view enables users and operators to interpret the rationale behind the alert in context. High-confidence episodes appear as long segments where $p_t$ remains consistently high with low variance, indicating stable slug behavior that is likely actionable. By contrast, borderline episodes are characterized by oscillations of $p_t$ around the decision threshold τ. Such cases are flagged for closer inspection and may suggest the need for manual review, threshold tuning, or adjustment of the persistence duration $\Delta_{min}$ his transparency builds operator trust and provides a principled way to refine alerting logic. For example, during start-up sequences where noisy transients are common, operators may raise $\Delta_{min}$ reduce false positives. Confidence traces thus bridge the gap between automated classification and operator-driven decision-making.

### 2.8 Snapshot Persistence for Reproducible Comparisons

Another important mechanism is snapshot persistence, designed to preserve results across reruns so that experiments remain comparable without retraining. In this approach, outputs such as evaluation & confusion matrices and thresholds are cached under a unique experiment key. The key is defined as, key = hash(model, hyperparameters, random seed, $\mathcal{T}_{train}, \mathcal{T}_{test}, \tau^*$). This ensures that results are tied to the exact model configuration, hyperparameters, data intervals, and decision threshold. By doing so, side-by-side comparisons of Decision Trees, Random Forests, and XGBoost models become possible with identical data slices, enhancing both reproducibility and auditability. In practice, only derived artifacts are stored—no raw confidential data—along with metadata such as software versions and experiment manifests. Thus, together, confidence trace visualization and snapshot persistence reinforce the reliability and interpretability of the workflow.

By embedding temporal integrity, event-level evaluation, persistence-based alerting, confidence assessment, and reproducible experiment management, the system transforms a standard machine learning pipeline into a decision-support tool that is aligned with industrial needs.

### 3. Dataset and Setup

The demonstration is tested using a simulated time-series dataset of oil well operations, spanning four months. The dataset captures key operational variables including inlet and outlet pressure, valve opening percentage, water levels in vessels, gas flow summation, and differential pressure. Each record is labeled as either *slug* or *non-slug*, enabling supervised training and evaluation of learners.

### 4. The Demonstration Tool

The system is delivered as an interactive web application that emphasizes usability, interpretability, and reproducibility, ensuring value both as a demonstration platform and as a practical deployment tool.

#### 4.1 Data ingestion and exploration

Users begin by uploading multivariate time-series measurements (e.g., inlet and outlet pressure, valve opening, vessel water level, differential pressure) in CSV format. Uploaded data can be previewed through expandable tables and filtered by time intervals.

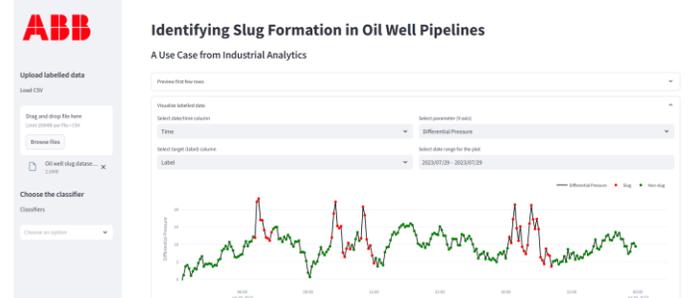

**Figure 2.** Dataset upload, preview & exploration

A built-in visualization module plots selected parameters against time, with slug and non-slug events highlighted by distinct markers (See Figure 2). This immediate visual feedback supports quick inspection of data quality and class balance.

#### 4.2 Label refinement

For datasets lacking annotations, the system allows interactive interval selection and manual labelling. This capability enables domain experts to refine annotations in alignment with operational knowledge, transforming raw sensor logs into machine-learning–ready training datasets (See Figure 3).

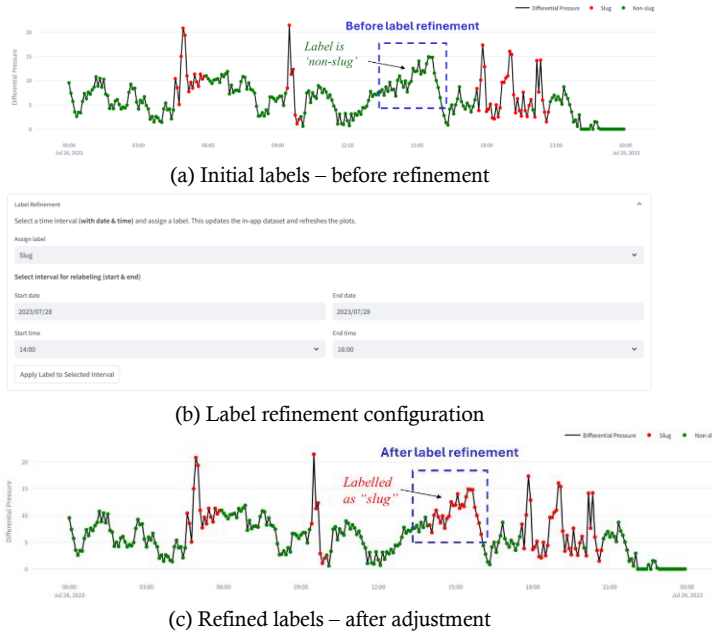

(a) Initial labels – before refinement

(b) Label refinement configuration

(c) Refined labels – after adjustment

**Figure 3.** Step-by-step process of refining labels

### 4.3 Model training and evaluation

Once labels are finalized, users can select from decision tree, random forest, or XGBoost classifiers, with configurable settings such as train–test splits (See Figure 4).

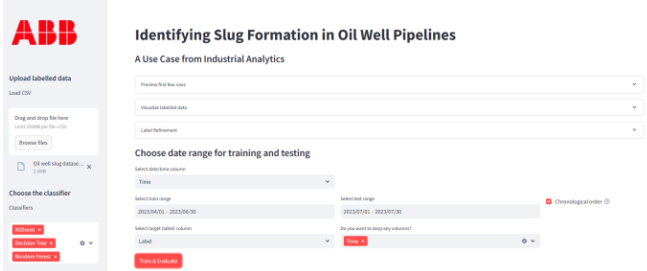

(a) Model training interface

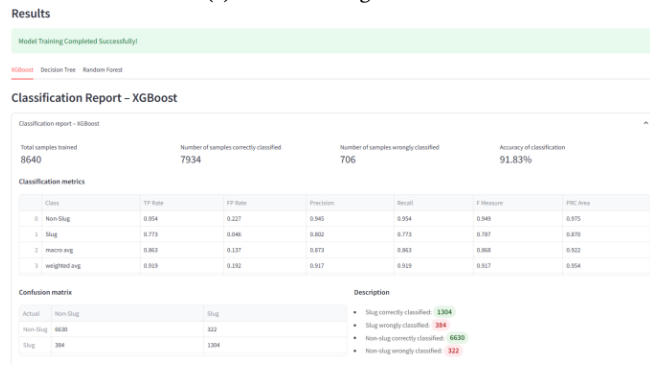

(b) Model evaluation results

**Figure 4.** Model training and evaluation workflow

Training progress is shown in real time, and results are immediately summarized through metrics and visual aids. A distinctive feature is snapshot persistence, which retains results across runs and allows side-by-side model comparisons without overwriting previous outcomes.

### 4.4 Visualization of results

Beyond aggregate metrics, predictions are projected onto the time axis, with explicit markers for true positives, false positives, true negatives, and false negatives (See Figure 5). These fine-grained visualizations allow users to examine the conditions under which misclassifications arise, a critical aspect in safety-sensitive domains where missed slug events (false negatives) carry significant risks.

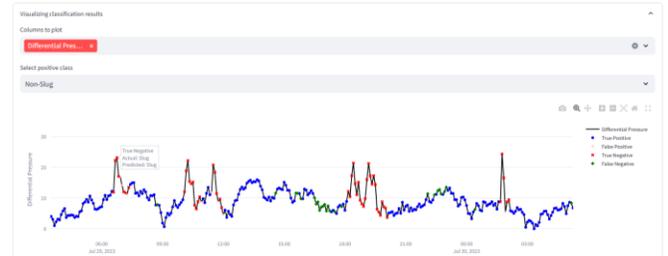

**Figure 5.** Visualization of results on test dataset

### 4.5 Inference and alert

For deployment-oriented evaluation, the inference module processes unlabeled or "real-time" datasets under conditions similar to live operations. The trained model outputs predictions with associated probability scores, which are visualized alongside sensor traces to provide context for decision-making. To suppress false alarms, a persistence-based alerting mechanism is applied, where an alert is raised only if slug predictions continue beyond a configurable temporal threshold. This approach balances sensitivity with robustness, ensuring that notifications correspond to sustained anomalies rather than brief transients (See Figure 6).

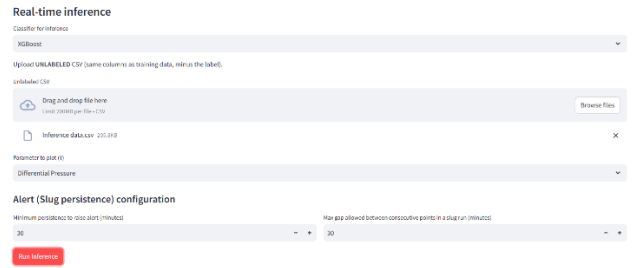

(a) Inference setup configuration

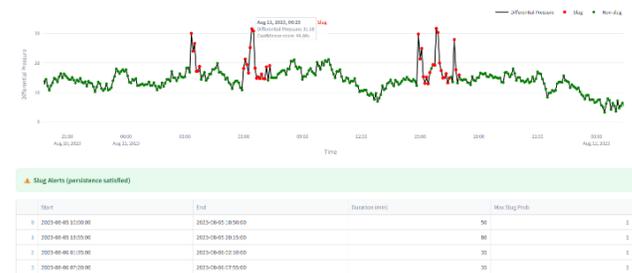

(b) Real-time inference output showing time-series classification, alerts, and persistence table

**Figure 6.** Real-time inference and alert generation

Overall, this tool highlights how ML–driven slug detection can be delivered not merely as an offline algorithm, but as a practical decision-support tool aligned with industrial operations.

## 5. Summary and future directions

The proposed system demonstrates how machine learning workflows can be reimagined as operator-facing tools rather than back-end algorithms. By emphasizing chronological integrity, event-focused evaluation, and transparent alerting, it highlights design choices that are critical for industrial acceptance but are often overlooked in purely academic studies. The interactive dashboard further illustrates the value of combining domain expertise with automated analytics, enabling users to engage directly with data, refine annotations, and interpret results in context. Beyond technical performance, the work shows that usability, reproducibility, and trustworthiness are equally important for transforming research into actionable solutions.

Looking ahead, the work will focus on enhancing the system's adaptability and real-time capabilities. Automated control feedback loops can be integrated to dynamically adjust flow rates, pump pressures, and valve positions in response to detected slug patterns. Transfer learning approaches will be explored to enable model reuse across different wells with minimal retraining. Incorporating fluid dynamics principles into machine learning algorithms can improve interpretability, while hybrid models linking data-driven learning with first-principles simulations may provide more reliable insights. Additionally, deploying edge AI models on low-power devices near wellheads and leveraging cloud-based analytics pipelines will support low-latency detection, large-scale monitoring, and coordinated anomaly management across multiple assets.

**Statement of Competing Interests:** The authors declare that they have no competing interests.

**Declaration of Generative AI and AI-assisted Technologies Use**: The authors declare that generative AI and AI-assisted technologies were used in the preparation of this work solely to improve readability and language. All content was generated under human oversight and has been thoroughly reviewed and edited by the authors. The authors remain fully responsible and accountable for the integrity, accuracy, and originality of the manuscript.